\documentclass[twoside,twocolumn,english]{IEEEtran}
\usepackage[T1]{fontenc}
\usepackage[latin9]{inputenc}
\usepackage{color}
\usepackage{babel}
\usepackage{refstyle}
\usepackage{amstext}
\usepackage{graphicx}
\usepackage{subscript}
\usepackage[unicode=true,pdfusetitle,
 bookmarks=true,bookmarksnumbered=false,bookmarksopen=false,
 breaklinks=false,pdfborder={0 0 1},backref=false,colorlinks=true]
 {hyperref}

\makeatletter

\AtBeginDocument{\providecommand\tabref[1]{\ref{tab:#1}}}
\AtBeginDocument{\providecommand\figref[1]{\ref{fig:#1}}}
\providecommand{\tabularnewline}{\\}
\newcommand{\lyxdot}{.}

\RS@ifundefined{subsecref}
  {\newref{subsec}{name = \RSsectxt}}
  {}
\RS@ifundefined{thmref}
  {\def\RSthmtxt{theorem~}\newref{thm}{name = \RSthmtxt}}
  {}
\RS@ifundefined{lemref}
  {\def\RSlemtxt{lemma~}\newref{lem}{name = \RSlemtxt}}
  {}

\usepackage{graphicx}

\makeatother

\begin{document}

\title{Facial Key Points Detection using Deep Convolutional Neural Network
- NaimishNet}

\author{Naimish Agarwal, IIIT-Allahabad (\textit{irm2013013@iiita.ac.in})\\
Artus Krohn-Grimberghe, University of Paderborn (\textit{artus@aisbi.de})\\
Ranjana Vyas, IIIT-Allahabad (\textit{ranjana@iiita.ac.in)}}
\maketitle
\begin{abstract}
Facial Key Points (FKPs) Detection is an important and challenging
problem in the fields of computer vision and machine learning. It
involves predicting the co-ordinates of the FKPs, e.g. nose tip, center
of eyes, etc, for a given face. In this paper, we propose a LeNet
adapted Deep CNN model - NaimishNet, to operate on facial key points
data and compare our model's performance against existing state of
the art approaches.
\end{abstract}

\begin{IEEEkeywords}
Facial Key Points Detection, Deep Convolutional Neural Network, NaimishNet
\end{IEEEkeywords}

\section{Introduction \label{sec:Introduction}}

Facial Key Points (FKPs) detection is an important and challenging
problem in the field of computer vision, which involves detecting
FKPs like centers and corners of eyes, nose tip, etc. The problem
is to predict the $\left(x,y\right)$ real-valued co-ordinates in
the space of image pixels of the FKPs for a given face image. It finds
its application in tracking faces in images and videos, analysis of
facial expressions, detection of dysmorphic facial signs for medical
diagnosis, face recognition, etc. 

Facial features vary greatly from one individual to another, and even
for a single individual there is a large amount of variation due to
pose, size, position, etc. The problem becomes even more challenging
when the face images are taken under different illumination conditions,
viewing angles, etc. 

In the past few years, advancements in FKPs detection are made by
the application of deep convolutional neural network (DCNN), which
is a special type of feed-forward neural network with shared weights
and local connectivity. DCNNs have helped build state-of-the-art models
for image recognition, recommender systems, natural language processing,
etc. Krizhevsky et al. \cite{key-1} applied DCNN in ImageNet image
classification challenge and outperformed the previous state-of-the-art
model for image classification. 

Wang et al. \cite{key-2} addressed FKPs detection by first applying
histogram stretching for image contrast enhancement, followed by principal
component analysis for noise reduction and mean patch search algorithm
with correlation scoring and mutual information scoring for predicting
left and right eye centers. Sun et al. \cite{key-3} estimated FKPs
by using a three level convolutional neural network, where at each
level, outputs of multiple networks were fused for robust and accurate
estimation. Longpre et al. \cite{key-4} predicted FKPs by first applying
data augmentation to expand the number of training examples, followed
by testing different architectures of convolutional neural networks
like LeNet \cite{key-5} and VGGNet \cite{key-6}, and finally used
a weighted ensemble of models. Nouri et al. \cite{key-7} used six
specialist DCNNs trained over pre-trained weights. Oneto et al. \cite{key-8}
applied a variety of data pre-processing techniques like histogram
stretching, Gaussian blurring, followed by image flipping, key point
grouping, and then finally applied LeNet. 

Taigman et al. \cite{key-9} provided a new deep network architecture,
DeepFace, for state-of- the-art face recognition. Li et al. \cite{key-10}
provided a new DCNN architecture for state-of-the art face alignment. 

We present a DCNN architecture \textendash{} NaimishNet, based on
LeNet, which addresses the problem of facial key points detection
by providing a learning model for a single facial key point. 

\section{NaimishNet Architecture \label{sec:NaimishNet-Architecture}}

NaimishNet consists of 4 convolution2d layers, 4 maxpooling2d layers
and 3 dense layers, with sandwiched dropout and activation layers,
as shown in \tabref{NaimishNet-layer-wise-architectu}.

\begin{table}
\begin{tabular}{c|c|c}
Layer Number & Layer Name & Layer Shape\tabularnewline
\hline 
1 & Input\textsubscript{1} & $\left(1,96,96\right)$\tabularnewline
2 & Convolution2d\textsubscript{1}  & $\left(32,93,93\right)$\tabularnewline
3 & Activation\textsubscript{1} & $\left(32,93,93\right)$\tabularnewline
4 & Maxpooling2d\textsubscript{1} & $\left(32,46,46\right)$\tabularnewline
5 & Dropout\textsubscript{1}  & $\left(32,46,46\right)$\tabularnewline
6 & Convolution2d\textsubscript{2} & $\left(64,44,44\right)$\tabularnewline
7 & Activation\textsubscript{2} & $\left(64,44,44\right)$\tabularnewline
8 & Maxpooling2d\textsubscript{2} & $\left(64,22,22\right)$\tabularnewline
9 & Dropout\textsubscript{2} & $\left(64,22,22\right)$\tabularnewline
10 & Convolution2d\textsubscript{3} & $\left(128,21,21\right)$\tabularnewline
11 & Activation\textsubscript{3} & $\left(128,21,21\right)$\tabularnewline
12 & Maxpooling2d\textsubscript{3} & $\left(128,10,10\right)$\tabularnewline
13 & Dropout\textsubscript{3} & $\left(128,10,10\right)$\tabularnewline
14 & Convolution2d\textsubscript{4} & $\left(256,10,10\right)$\tabularnewline
15 & Activation\textsubscript{4} & $\left(256,10,10\right)$\tabularnewline
16 & Maxpooling2d\textsubscript{4} & $\left(256,5,5\right)$\tabularnewline
17 & Dropout\textsubscript{4} & $\left(256,5,5\right)$\tabularnewline
18 & Flatten\textsubscript{1} & $\left(6400\right)$\tabularnewline
19 & Dense\textsubscript{1} & $\left(1000\right)$\tabularnewline
20 & Activation\textsubscript{5} & $\left(1000\right)$\tabularnewline
21 & Dropout\textsubscript{5} & $\left(1000\right)$\tabularnewline
22 & Dense\textsubscript{2} & $\left(1000\right)$\tabularnewline
23 & Activation\textsubscript{6} & $\left(1000\right)$\tabularnewline
24 & Dropout\textsubscript{6} & $\left(1000\right)$\tabularnewline
25 & Dense\textsubscript{3} & $\left(2\right)$\tabularnewline
\end{tabular}\caption{NaimishNet layer-wise architecture \label{tab:NaimishNet-layer-wise-architectu}}
\end{table}

\subsection{Layer-wise Details}

The following points give the details of every layer in the architecture:
\begin{itemize}
\item Input\textsubscript{1 }is the input layer.
\item Activation\textsubscript{1} to Activation\textsubscript{5} use Exponential
Linear Units (ELUs) \cite{key-11} as activation functions, whereas
Activation\textsubscript{6} uses Linear Activation Function. 
\item Dropout \cite{key-12} probability is increased from $0.1$ to $0.6$
from Dropout\textsubscript{1} to Dropout\textsubscript{6}, with
a step size of $0.1$. 
\item Maxpooling2d\textsubscript{1} to Maxpooling2d\textsubscript{4} use
a pool shape of $(2,2)$, with non-overlapping strides and no zero
padding. 
\item Flatten\textsubscript{1} flattens 3d input to 1d output. 
\item Convolution2d\textsubscript{1} to Convolution2d\textsubscript{4}
do not use zero padding, have their weights initialized with random
numbers drawn from uniform distribution, and the specifics of number
of filters and filter shape are shown in \tabref{Filter-details-of}.
\item Dense\textsubscript{1} to Dense\textsubscript{3} are regular fully
connected layers with weights initialized using Glorot uniform initialization
\cite{key-13}. 
\item Adam \cite{key-14} optimizer, with learning rate of $0.001$, $\beta_{1}$
of $0.9$, $\beta_{2}$ of $0.999$ and $\varepsilon$ of $1e-08$,
is used for minimizing Mean Squared Error (MSE).
\item There are $7,488,962$ model parameters. 
\end{itemize}
\begin{table}
\begin{tabular}{c|c|c}
Layer Name & Number of Filters & Filter Shape\tabularnewline
\hline 
Convolution2d\textsubscript{1} & $32$ & $\left(4,4\right)$\tabularnewline
Convolution2d\textsubscript{2} & $64$ & $\left(3,3\right)$\tabularnewline
Convolution2d\textsubscript{3} & $128$ & $\left(2,2\right)$\tabularnewline
Convolution2d\textsubscript{4} & $256$ & $\left(1,1\right)$\tabularnewline
\end{tabular}

\caption{Filter details of Convolution2d layers\label{tab:Filter-details-of}}
\end{table}

\subsection{Design Decisions}

We have tried various different deep network architectures before
finally settling down with NaimishNet architecture. Most of the design
decisions were governed primarily by the hardware restrictions and
the time taken per epoch for different runs of the network. Although
we wanted one architecture to fit well to all the different FKPs as
target, we have mainly taken decisions based on the performance of
the model on left eye center FKP as target. We have run the different
architectures for at most $40$ epochs and forecasted its performance
and generalizability based on our experience we gathered while training
multiple deep network architectures. 

We experimented with 5 convolution2d and 5 maxpooling2d layers and
4 dense layers with rectified linear units as activation functions,
but that over fitted. We experimented with 3 convolution2d and 3 maxpooling2d
layers and 2 dense layers with rectified linear units as activation
functions, but that in turn required pre-training the weights to achieve
good performance. We wanted to avoid pre-training to reduce the training
time. We did not experiment with zero padding and strides. 

We sandwiched batch normalization layers \cite{key-15} after every
maxpooling2d layer but that increased the per epoch training time
by fivefold. We used dropout layers with dropout probability of 0.5
for every dropout layer but that showed a poor fit. We removed the
dropout layers but that resulted in overfitting. We tried different
initialization schemes like He normal initialization \cite{key-15},
but that did not show nice performance in the first 20 epochs, in
fact uniform initialization performed better than that for us. We
experimented with different patience levels (PLs) for early stopping
callback function and concluded that PL of 10\% of total number of
epochs, in our case 300, would do well in most of the cases. For the
aforementioned cases, we used RMSProp optimizer \cite{key-16}. 

The main aim was to reduce the training time along with achieving
maximum performance and generalizability. To achieve that we experimented
with exponential linear units as activation function, Adam optimizer,
and linearly increasing dropout probabilities along the depth of the
network. The specifics of the filter were chosen based on our past
experience with exponentially increasing number of filters and linearly
decreasing filter size. The batch size was kept 128 because vector
operations are optimized for sizes which are in powers of 2. Also,
given over 7 million model parameters, and limited GPU memory, batch
size of 128 was an ideal size for us. The number of main layers i.e.
4 convolutional layers (convolution2d and maxpooling2d) and 3 dense
layers was ideal since exponential linear units perform well in cases
where the number of layers are greater than 5. The parameters of Adam
optimizer were used as is as recommended by its authors and we did
not experiment with them. 

NaimishNet, is inspired by LeNet \cite{key-5}, since LeNet\textquoteright s
architecture follows the pattern \emph{input $\rightarrow$ conv2d
$\rightarrow$ maxpool2d $\rightarrow$ conv2d $\rightarrow$ maxpool2d
$\rightarrow$ conv2d $\rightarrow$ maxpool2d $\rightarrow$ dense
$\rightarrow$ output.} Works of Longpre et al. \cite{key-4}, Oneto
et al. \cite{key-8}, etc. have shown that LeNet styled architectures
have been successfully applied for Facial Key Points Detection problem. 

\section{Experiments \label{sec:Experiments}}

The details of the experiment are given below.

\subsection{Dataset Description}

We have used the dataset from Kaggle Competition \textendash{} Facial
Key Points Detection \cite{key-17}. The dataset was chosen to benchmark
our solution against the existing approaches which address FKPs detection
problem. 

There are 15 FKPs per face image like left eye center, right eye center,
left eye inner corner, left eye outer corner, right eye inner corner,
right eye outer corner, left eyebrow inner end, left eyebrow outer
end, right eyebrow inner end, right eyebrow outer end, nose tip, mouth
left corner, mouth right corner, mouth center top lip and mouth center
bottom lip. Here, left and right are from the point of view of the
subject. 

The greyscale input images, with pixel values in range of $[0,255]$,
have size of $96\times96$ pixels. The train dataset consists of 7049
images, with 30 targets, i.e. $(x,y)$ for each of 15 FKPs. It has
missing target values for many FKPs for many face images. The test
dataset consists of 1783 images with no target information. The Kaggle
submission file consists of 27124 FKPs co-ordinates, which are to
be predicted. 

The Kaggle submissions are scored based on the Root Mean Squared Error
(RMSE), which is given by 
\[
RMSE=\sqrt{\frac{\sum_{i=1}^{n}\left(y_{i}-\hat{y_{i}}\right)^{2}}{n}}
\]

where $y_{i}$ is the original value, $\hat{y}_{i}$ is the predicted
value and $n$ is the number of targets to be predicted. 

\subsection{Proposed Approach}

We propose the following approach for Facial Key Points Detection.

\subsubsection{Data Augmentation}

Data Augmentation helps boost the performance of a deep learning model
when the training data is limited by generating more training data.
We have horizontally flipped \cite{key-7} the images for which target
information for all the 15 FKPs are available, and also swapped the
target values according to \tabref{Target-Facial-Key}. Then, we vertically
stacked the new horizontally flipped data under the original train
data to create the augmented train dataset. 

\begin{table}
\resizebox{0.49\textwidth}{!}{%

\begin{tabular}{c|c}
Target Facial Key Points & Target Facial Key Points\tabularnewline
\hline 
Left Eye Center X  & Right Eye Center X \tabularnewline
Left Eye Center Y  & Right Eye Center Y \tabularnewline
Left Eye Inner Corner X  & Right Eye Inner Corner X \tabularnewline
Left Eye Inner Corner Y  & Right Eye Inner Corner Y \tabularnewline
Left Eye Outer Corner X  & Right Eye Outer Corner X \tabularnewline
Left Eye Outer Corner Y  & Right Eye Outer Corner Y \tabularnewline
Left Eyebrow Inner Corner X  & Right Eyebrow Inner Corner X \tabularnewline
Left Eyebrow Inner Corner Y  & Right Eyebrow Inner Corner Y \tabularnewline
Left Eyebrow Outer Corner X  & Right Eyebrow Outer Corner X \tabularnewline
Left Eyebrow Outer Corner Y  & Right Eyebrow Outer Corner Y \tabularnewline
Mouth Left Corner X  & Mouth Right Corner X \tabularnewline
Mouth Left Corner Y  & Mouth Right Corner Y \tabularnewline
\end{tabular}}

\caption{Target Facial Key Points which need to be swapped while horizontally
flipping the data \label{tab:Target-Facial-Key}}
\end{table}

\subsubsection{Data Pre-processing}

The image pixels are normalized to the range $[0,1]$ by dividing
by 255.0, and the train targets are zero-centered to the range $[-1,1]$
by first dividing by 48.0, since the images are $96\times96$, and
then subtracting 48.0. 

\subsubsection{Pre-training Analysis}

\begin{figure}
\includegraphics[scale=0.55]{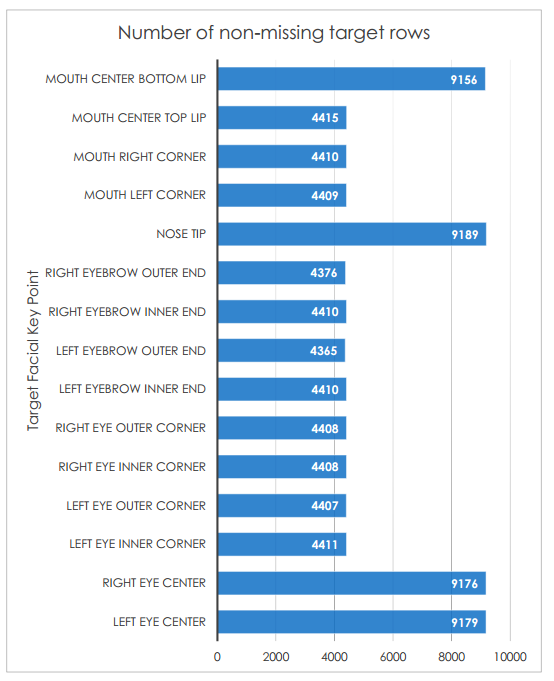}

\caption{Number of non-missing target rows in the augmented and pre-processed
train dataset \label{fig:Number-of-non-missing}}
\end{figure}

We have created 15 NaimishNet models, since \figref{Number-of-non-missing}
shows that there are different number of non-missing target rows for
different FKPs. 

\subsubsection{Training}

For each of the 15 NaimishNet models, filter (keep) the rows with
non-missing target values and split the filtered train data into 80\%
train dataset (T) and 20\% validation dataset (V). Compute feature-wise
mean M on T, and zero-center T and V by subtracting M, and also store
M. Reshape features of T and V to $(1,96,96)$, and start the training.
Finally, validate on V, and store the model\textquoteright s loss
history. 

Each model is trained with a batch size of 128 and maximum number
of epochs are 300. Two callback functions are used which execute at
the end of every epoch \textendash{} Early Stopping Callback (ESC)
and Model Checkpoint Callback (MCC). 

ESC stops the training when the number of contiguous epochs without
improvement in validation loss are 30 (Patience Level). MCC stores
the weights of the model with the lowest validation loss. 

\subsubsection{Evaluation}

For each of the 15 NaimishNet models, reload M, zero center test features
by subtracting M and reshape the samples to $(1,96,96)$. Reload the
best check-pointed model weights, predict on test features, and store
the predictions. 

\section{Results \label{sec:Results}}

\begin{figure}
\includegraphics[scale=0.4]{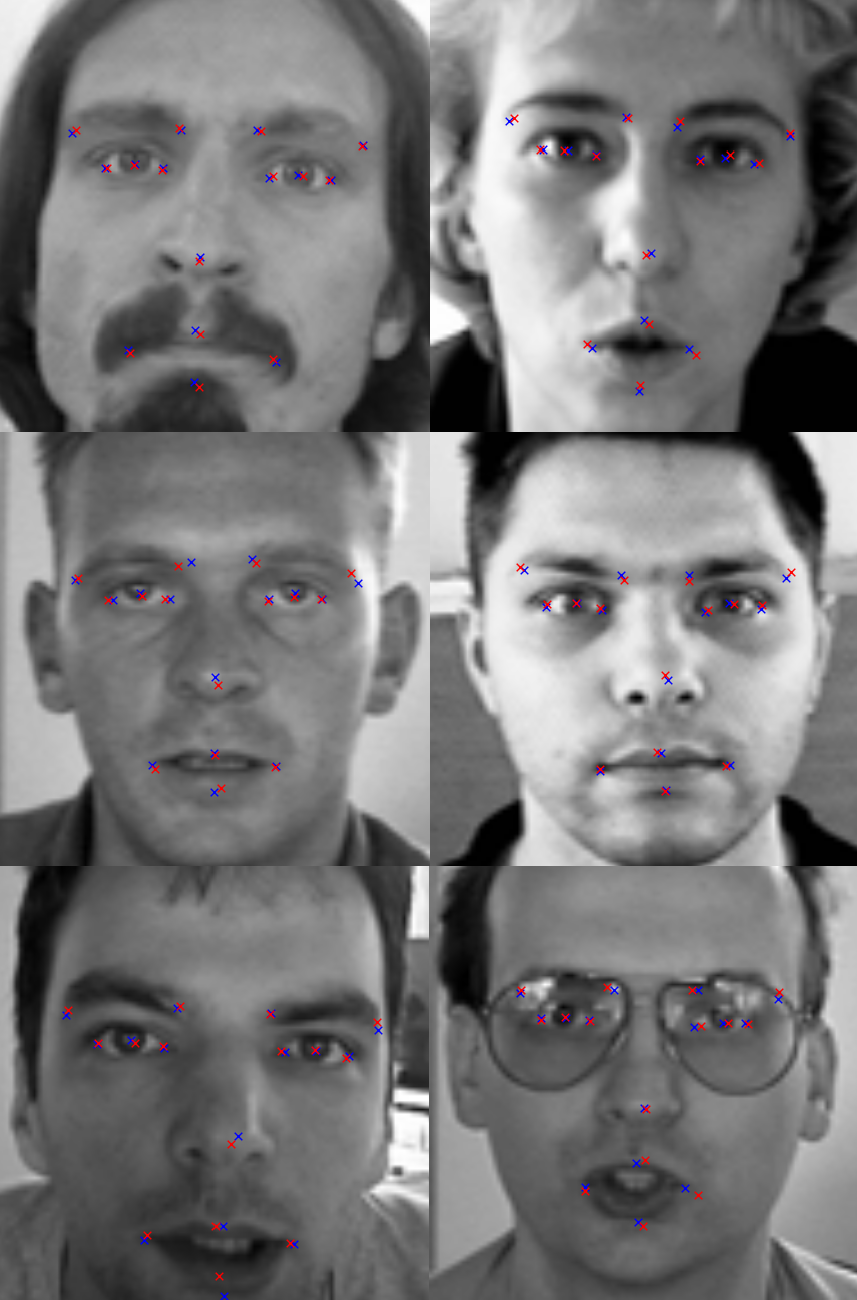}

\caption{Annotated Faces with 15 Facial Key Points, marked in blue for original
and in red for predicted. \label{fig:Annotated-Faces-with}}
\end{figure}

The annotated faces with 15 FKPs of 6 individuals is shown in \figref{Annotated-Faces-with}.
It is visible that the predicted locations are very close to the actual
locations of FKPs. 

In figure \ref{fig:The-worst,-medium}, high RMSE\textsubscript{left}
is because of the presence of less number of images, in the training
dataset, with the side face pose. Most of the images in the training
dataset are that of the frontal face with nearly closed lips. So,
NaimishNet delivers lower RMSE\textsubscript{center} and RMSE\textsubscript{right}. 

\begin{figure}
\includegraphics[scale=0.55]{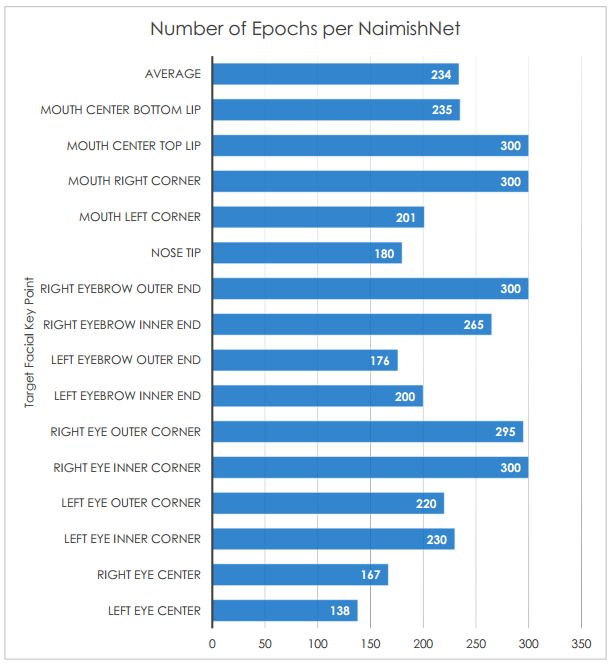}

\caption{Number of Epochs per NaimishNet \label{fig:Number-of-Epochs}}
\end{figure}

Total number of epochs are 3507, as shown in \figref{Number-of-Epochs}.
Total training time of 15 NaimishNet models is approximately 20 hours. 

\begin{figure}

\includegraphics[scale=0.7]{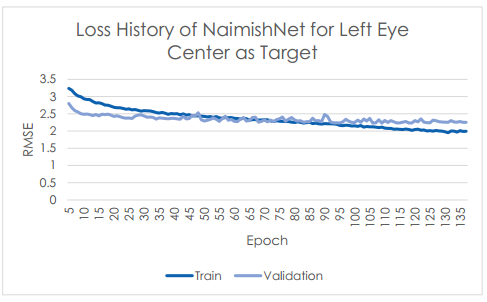}

\caption{Loss History of NaimishNet for Left Eye Center as Target \label{fig:left-eye-center}}
\end{figure}

\begin{figure}
\includegraphics[scale=0.7]{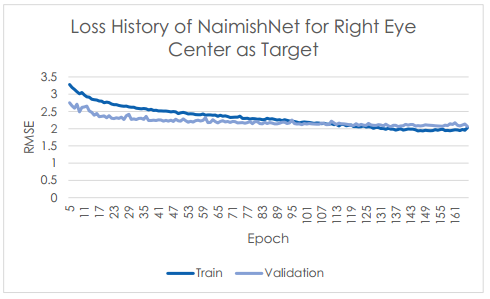}

\caption{Loss History of NaimishNet for Right Eye Center as Target \label{fig:right-eye-center}}
\end{figure}

\begin{figure}

\includegraphics[scale=0.7]{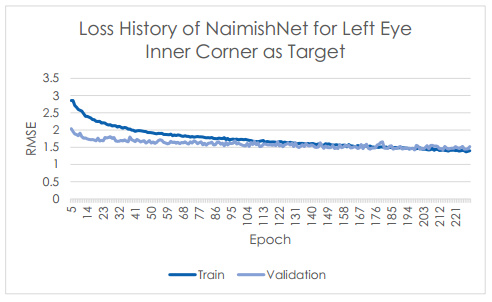}

\caption{Loss History of NaimishNet for Left Eye Inner Corner as Target \label{fig:left-eye-inner-corner}}

\end{figure}

\begin{figure}

\includegraphics[scale=0.7]{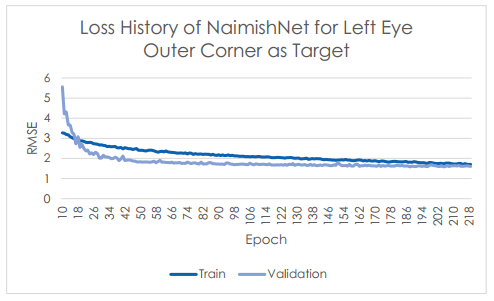}

\caption{Loss History of NaimishNet for Left Eye Outer Corner as Target \label{fig:left-eye-outer-corner}}

\end{figure}

\begin{figure}

\includegraphics[scale=0.7]{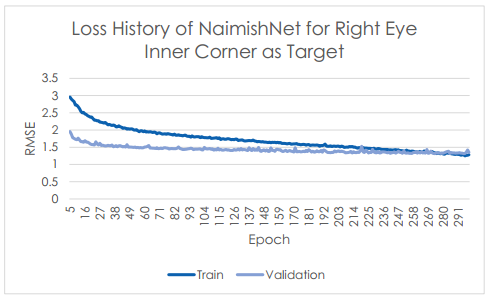}

\caption{Loss History of NaimishNet for Right Eye Inner Corner as Target \label{fig:right-eye-inner-corner}}

\end{figure}

\begin{figure}

\includegraphics[scale=0.7]{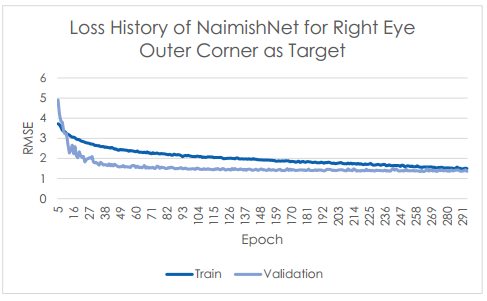}

\caption{Loss History of NaimishNet for Right Eye Outer Corner as Target \label{fig:right-eye-outer-corner}}
\end{figure}

\begin{figure}

\includegraphics[scale=0.7]{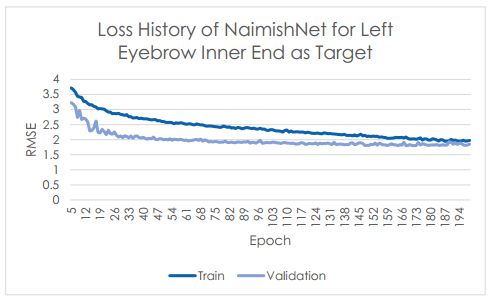}

\caption{Loss History of NaimishNet for Left Eyebrow Inner End as Target \label{fig:left-eyebrow-inner-end}}

\end{figure}

\begin{figure}

\includegraphics[scale=0.7]{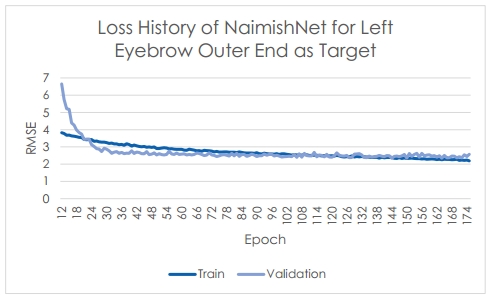}

\caption{Loss History of NaimishNet for Left Eyebrow Outer End as Target \label{fig:left-eyebrow-outer-end}}

\end{figure}

\begin{figure}

\includegraphics[scale=0.7]{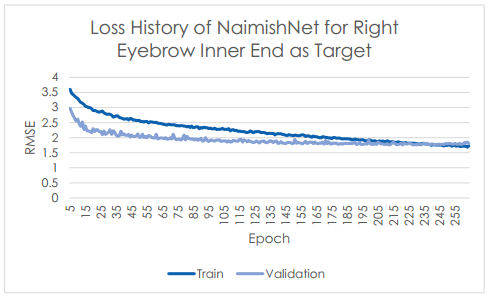}

\caption{Loss History of NaimishNet for Right Eyebrow Inner End as Target \label{fig:right-eyebrow-inner-end}}

\end{figure}

\begin{figure}

\includegraphics[scale=0.7]{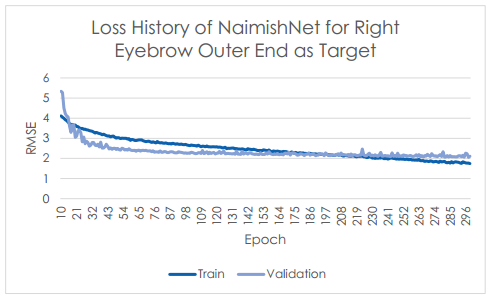}

\caption{Loss History of NaimishNet for Right Eyebrow Outer End as Target \label{fig:right-eyebrow-outer-end}}
\end{figure}

\begin{figure}
\includegraphics[scale=0.7]{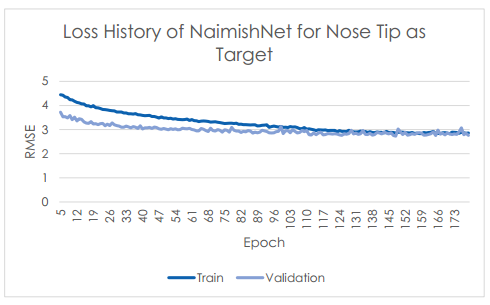}

\caption{Loss History of NaimishNet for Nose Tip as Target \label{fig:nose-tip}}

\end{figure}

\begin{figure}

\includegraphics[scale=0.7]{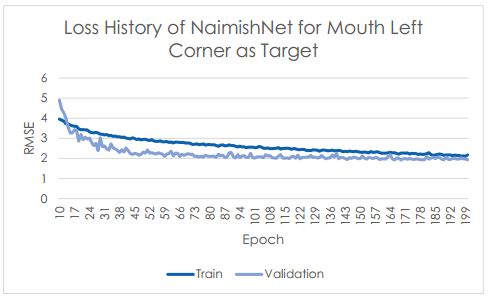}

\caption{Loss History of NaimishNet for Mouth Left Corner as Target \label{fig:mouth-left-corner}}

\end{figure}

\begin{figure}

\includegraphics[scale=0.7]{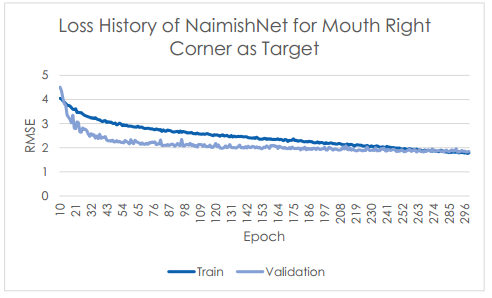}

\caption{Loss History of NaimishNet for Mouth Right Corner as Target \label{fig:mouth-right-corner}}

\end{figure}

\begin{figure}

\includegraphics[scale=0.7]{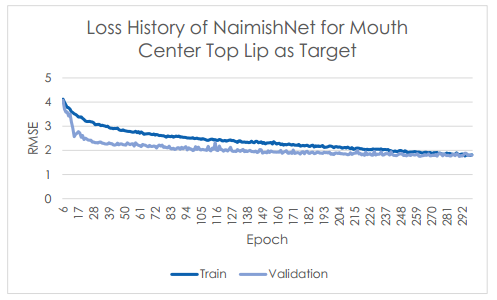}

\caption{Loss History of NaimishNet for Mouth Center Top Lip as Target \label{fig:mouth-center-top-lip}}

\end{figure}

\begin{figure}

\includegraphics[scale=0.7]{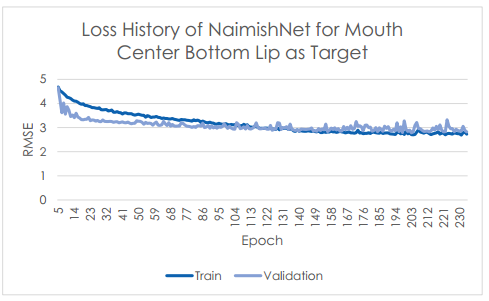}

\caption{Loss History of NaimishNet for Mouth Center Bottom Lip as Target \label{fig:mouth-center-bottom-lip} }

\end{figure}

\begin{figure}

\includegraphics[scale=0.7]{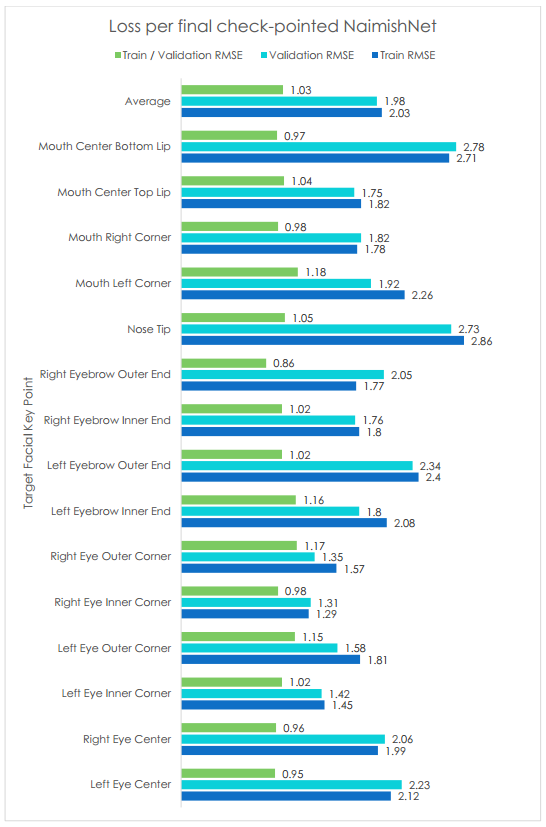}

\caption{Loss per final check-pointed NaimishNet \label{fig:Loss-per-final}}

\end{figure}

In the following points, we present the analysis of train and validation
error for each of the 15 NaimishNet models taken 2 at a time:
\begin{itemize}
\item Both \figref{left-eye-center} and \figref{right-eye-center} show
that both train and validation RMSE decrease over time with small
wiggles along the curve showcasing that the batch size was rightly
chosen and end up being close enough, thus showing that the two NaimishNet
models generalized well. The best check-pointed models were achieved
before 300 epochs, thus certifying the decision of fixing 300 epochs
as the upper limit. 
\item Both \figref{left-eye-inner-corner} and \figref{left-eye-outer-corner}
show that both train and validation RMSE decrease over time with small
wiggles and end up being close enough. The best check-pointed models
were achieved before 300 epochs. Also, for \figref{left-eye-outer-corner},
since the train RMSE does not become less than validation RMSE implies
that the Patience Level (PL) for ESC could have been increased to
50. Thus, the training could have been continued for some more epochs. 
\item Both \figref{right-eye-inner-corner} and \figref{right-eye-outer-corner}
show that both train and validation RMSE decrease over time with very
small wiggles and end up being close enough. The best check-pointed
models were achieved close to 300 epochs. For \figref{right-eye-outer-corner},
PL of ESC could have been increased to 50 so that the train RMSE becomes
less than validation RMSE. 
\item Both \figref{left-eyebrow-inner-end} and \figref{left-eyebrow-outer-end},
show that the train and validation RMSE decreases over time with small
wiggles, and end up being close enough. The best check-pointed models
were achieved far before 300 epochs. No further improvement could
have been seen by increasing PL of ESC. 
\item Both \figref{right-eyebrow-inner-end} and \figref{right-eyebrow-outer-end}
show that both train and validation RMSE decrease over time with small
wiggles and end up being close in the check-pointed model, which can
be seen 30 epochs before the last epoch on x-axis.
\item Both \figref{nose-tip} and \figref{mouth-left-corner} show that
both train and validation RMSE decrease over time and they end up
being close enough in the final check-pointed model. For \figref{mouth-left-corner},
PL of ESC could have been increased to 50 but not necessarily there
would have been any visible improvement. 
\item Both \figref{mouth-right-corner} and \figref{mouth-center-top-lip}
show that the train and validation RMSE decrease over time, with small
wiggles, and the upper limit of 300 epochs was just right enough for
the models to show convergence. 
\item As per \figref{mouth-center-bottom-lip}, the train and validation
RMSE decreases over time and the wiggly nature of the validation curve
increases towards the end. We could have tried to increase the batch
size to 256 if it could be handled by our hardware. The curves are
close together and show convergence before 300 epochs. 
\end{itemize}
As per \figref{Loss-per-final}, the NaimishNet generalizes well,
since the average train RMSE / validation RMSE is 1.03, where average
loss is calculated by 

\[
\text{Average RMSE = \ensuremath{\sqrt{\frac{\sum_{i=1}^{15}RMSE_{i}^{2}}{15}}}}
\]

\begin{figure}

\includegraphics[scale=0.7]{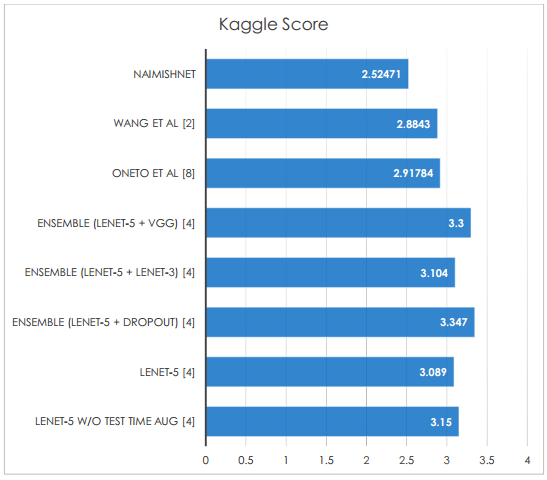}

\caption{Comparison of Kaggle Scores \label{fig:Comparison-of-Kaggle}}

\end{figure}

As per \figref{Comparison-of-Kaggle}, the NaimishNet outperforms
the approaches used by Longpre et al. \cite{key-4}, Oneto et al.
\cite{key-8}, Wang et al. \cite{key-2}, since lower the score, better
the model. 

\begin{figure}
\includegraphics[scale=0.25]{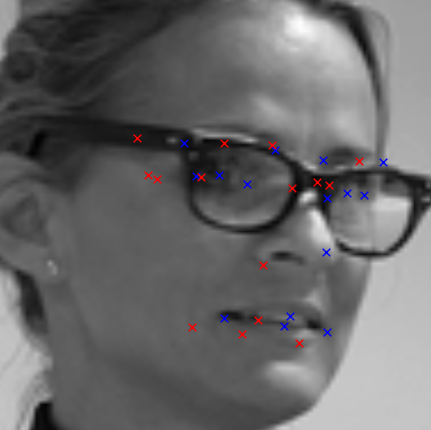}\includegraphics[scale=0.25]{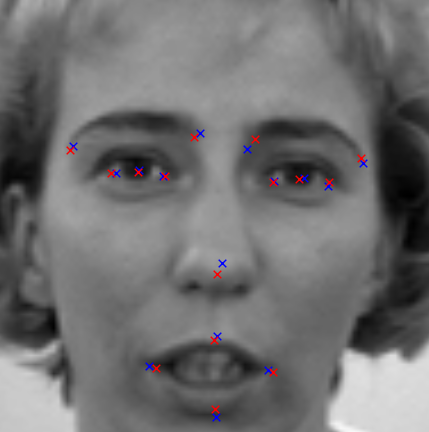}\includegraphics[scale=0.25]{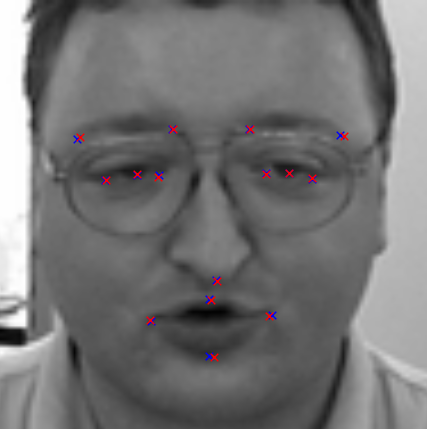}

\caption{The worst, medium and best predictions, where the RMSE\protect\textsubscript{left}
= 6.87, RMSE\protect\textsubscript{center} = 1.04, and RMSE\protect\textsubscript{right}
= 0.37, the original FKPs are shown in blue and the predicted FKPs
are shown in red. \label{fig:The-worst,-medium}}
\end{figure}

\section{Conclusion}

NaimishNet \textendash{} a learning model for a single facial key
point, is based on LeNet \cite{key-5}. As per \figref{Comparison-of-Kaggle},
LeNet based architectures have proven to be successful for Facial
Key Points Detection. Works of Longpre et al. \cite{key-4}, Oneto
et al. \cite{key-8}, etc. have shown that LeNet styled architectures
have been successfully applied for Facial Key Points Detection problem.

\section{Future Scope}

Given enough compute resources and time, NaimishNet can be further
modified by experimenting with different initialization schemes, different
activation functions, different number of filters and kernel size,
different number of layers, switching from LeNet \cite{key-5} to
VGGNet \cite{key-6}, introducing Inception modules as in GoogleNet
\cite{key-18}, or introducing Residual Networks \cite{key-19}, etc.
Different image pre-processing approaches like histogram stretching,
zero centering, etc. can be tried to check which approaches improve
the model\textquoteright s accuracy. New approaches to Facial Key
Points Detection can be based on the application of deep learning
as a three stage process - face detection, face alignment, and facial
key points detection, with the use of the state-of-the-art algorithms
for each stage.

\end{document}